\pdfoutput=1

\documentclass[11pt]{article}

\usepackage[preprint]{acl}

\usepackage{times}
\usepackage{latexsym}

\usepackage[T1]{fontenc}

\usepackage[utf8]{inputenc}

\usepackage{microtype}

\usepackage{inconsolata}

\usepackage{amsmath}

\title{Psychological Assessments with Large Language Models:\\A Privacy-Focused and Cost-Effective Approach}

\author{Sergi Blanco-Cuaresma \\
  Harvard-Smithsonian Center for Astrophysics, 60 Garden Street, Cambridge, MA 02138, USA \\
  Laboratoire de Recherche en Neuroimagerie, University Hospital (CHUV) and \\University of Lausanne (UNIL), Lausanne, Switzerland \\
  \texttt{sblancocuaresma@cfa.harvard.edu} \\\
}

\begin{document}
\maketitle
\begin{abstract}
This study explores the use of Large Language Models (LLMs) to analyze text comments from Reddit users, aiming to achieve two primary objectives: firstly, to pinpoint critical excerpts that support a predefined psychological assessment of suicidal risk; and secondly, to summarize the material to substantiate the preassigned suicidal risk level. The work is circumscribed to the use of "open-source" LLMs that can be run locally, thereby enhancing data privacy. Furthermore, it prioritizes models with low computational requirements, making it accessible to both individuals and institutions operating on limited computing budgets. The implemented strategy only relies on a carefully crafted prompt and a grammar to guide the LLM's text completion. Despite its simplicity, the evaluation metrics show outstanding results, making it a valuable privacy-focused and cost-effective approach. This work is part of the Computational Linguistics and Clinical Psychology (CLPsych) 2024 shared task.
\end{abstract}

\section{Introduction}

Large Language Models (LLMs) like GPT \citep[Generative Pre-trained Transformer;][]{2023arXiv230308774O}, Llama \citep[Large Language Model Meta AI;][]{2023arXiv230213971T, 2023arXiv230709288T}, Mistral/Mixtral \citep{2024arXiv240104088J, 2023arXiv231006825J}, and others \citep[based on the transformer architecture and its attention mechanism, made popular thanks to BERT and derivatives;][]{2017arXiv170603762V, 2018arXiv181004805D, 2021arXiv211200590G, 2022arXiv221200744G} represent a significant advancement in the field of artificial intelligence, specifically within natural language processing (NLP). These models have transformed how machines understand, generate, and interact with human language, enabling a wide range of applications.

During the "pre-training" phase, LLMs learn a wide range of language patterns and they encode knowledge from a vast corpora of text data. In a posterior phase, they can be "fine-tuned" on smaller/alternate datasets to become specialized on specific tasks such as psychological assessments. The fine-tuning can also be restricted to a smaller number of parameters using techniques such as LoRA \citep{2021arXiv210609685H} or QLoRA \citep{2023arXiv230514314D} for quantized models \citep{2023arXiv230214017K}. However, fine-tuning can still be costly in terms of computational resources and time investment, requiring a high level of expertise. 

Models with a higher number of parameters are more sophisticated, encode more accurate knowledge and are capable of performing more advanced tasks with optimal results. This reduces the need for fine-tuning, but it increases the requirements for computational resources. There is also the option of not running the models locally, but relying on external services such as OpenAI's API and their chatGPT interface\footnote{\url{https://openai.com/}}. Regrettably, this approach may not be viable due to the involvement of third parties, which might not ensure adequate data protection or adhere to the stringent privacy standards and ethical codes mandated by healthcare and medical institutions, along with other legal obligations.

Given this context, in this study I explore the use of "open-source" LLMs that can be run locally in current commodity hardware (thus, 4-bit quantized models with a maximum 7 billion parameters), and I do not fine-tune these models to specialize in any specific task or to incorporate new knowledge relevant to the domain of clinical psychology. This evaluation is focused on the shared task proposed by the Computational Linguistics and Clinical Psychology (CLPsych) 2024 workshop \citep{chim2024overview} at the 18th Conference of the European Chapter of the Association for Computational Linguistics (ACL).

\section{Task and Data}

The CLPsych 2024 shared task consisted on finding evidence within Reddit comments that support a preassigned suicide risk level. The organizers provided access the University of Maryland Reddit Suicidality Dataset \citep[UMD version 2;][]{shing-etal-2018-expert, zirikly-etal-2019-clpsych}, which includes posts to the \textit{"r/SuicideWatch"} subreddit plus crowdsourced and expert risk level assessments. The risk levels to be considered are low, moderate and high.

The evidence supporting the preassigned risk level can take two different forms: 1) highlights (i.e., snippets) from the user's comments; 2) a summary that aggregates the evidence that justifies the assigned risk level. In this study, both forms of evidence were generated for a selection of 162 posts (by 125 users) that the organizers used to evaluate each submitted result.

\section{Methods}

This study considered six different LLMs, which were selected based on their ranking on the Open LLM Leaderboard\footnote{\url{https://huggingface.co/spaces/HuggingFaceH4/open_llm_leaderboard}}, and the LMSys Chatbot Arena Leaderboard\footnote{\url{https://huggingface.co/spaces/lmsys/chatbot-arena-leaderboard}} as of January 15th (2024). The models were obtained from Tom Jobbins's huggingface repository\footnote{\url{https://huggingface.co/TheBloke}} in GGUFv2 format ("Q4\_K\_M" quant method). The inference code was run locally using the NASA SciX Brain software \citep{2023arXiv231214211B} on a MacBook Air with the Apple M1 chip (released on November 2020) and 16GB of RAM. The concrete models were:

\begin{enumerate}
    \item OpenHermes 2.5\footnote{\url{https://huggingface.co/teknium/OpenHermes-2.5-Mistral-7B}}, based on Mistral 7B and further trained on mainly GPT-4 generated data, and other open datasets.
    \item Orca 2 \citep{2023arXiv231111045M}, based on Llama 2, designed to excel in reasoning, trained on a censored synthetic dataset. Human preference alignment techniques such as Reinforcement learning from human feedback \citep[RLHF;][]{2019arXiv190908593Z} or Direct Preference Optimization \citep[DPO;][]{2023arXiv230518290R} were not used.
    \item Starling 7B alpha \citep{starling2023}, based on OpenChat 3.5 which is refinement of Mistral 7B using C(onditioned)-RLFT \citep{2023arXiv230911235W}, trained by Reinforcement Learning from AI Feedback \citep[RLAIF;][]{2023arXiv230900267L}.
    \item Dolphin 2.6, based on Mistral 7B, trained following LASER \citep{2023arXiv231213558S} and aligned to human preferences using DPO.
    \item Mistral 7B instruct 0.2 \citep{2024arXiv240104088J}, based on Mistral 7B, trained with a variety of publicly available conversation datasets.
    \item Zephyr 7B beta \citep{2023arXiv231016944T}, based on Mistral 7B, trained on on a mix of publicly available, synthetic datasets using DPO.
\end{enumerate}

Each model is requested to either extract evidences from user's comments as text highlights or to generate a comprehensive summary, both with the goal of justifying a preassigned suicidal risk level. The request is done with a crafted prompt that sets the scene (e.g., act as a psychologist specializing in suicidal ideation), and includes a fake interaction where the user has shared the reddit comment and a preassigned risk level, and the model (i.e., the assistant) has already provided an answer. This is a one-shot prompt from where the model can infer what we expect it to generate after a user request. Subsequently, the real comment to be analyzed is included, and the model's response is left empty for it to be completed.

My evaluation of various prompts was not exhaustive, but rather a manual and subjective process based on a limited set of examples. The tests \citep[inspired by a previous work;][]{2023arXiv231214211B} suggested that a prompt in which a user outlines the entire task and includes an example generally yields slightly inferior results compared to a prompt that simulates an initial round of interaction between the user and the assistant, as if the assistant had already responded to a previous request. All the tested prompts directed the model to adopt the role of an expert and incorporated a description of what constitutes evidence for supporting a suicidal risk assessment, based on the assumption that the LLM will rely more heavily on this provided information than on the knowledge it has gained through its training.

The structure of the final prompt used for extracting text highlights can be found in Appendix~\ref{sec:appendix_prompt_highlights}. For this particular subtask, I  use a formal grammar (feature included in llama.cpp\footnote{\url{https://github.com/ggerganov/llama.cpp}}) to constraint the possible tokens that can be sampled (i.e., discarding tokens that would break the rules defined by the grammar). The grammar is in GBNF format, which is an extension of BNF (Backus–Naur/Normal form, a metasyntax notation for context-free grammars) that primarily adds a few modern regex-like features. The grammar imposes the generation of a list (surrounded by square brackets) of strings (surrounded by double quotes), and the string can only contain words present in the user's comment in their original order (see a concrete example in Appendix~\ref{sec:appendix_grammar_highlights}).

Regarding the summarization subtask, the structure of the used prompt can be found in Appendix~\ref{sec:appendix_prompt_summary}. In this case, there is no imposed grammar, the model is free to complete the response but it is primed by providing already the first sentence, which states what the preassigned suicidal risk level is.

For both subtasks, a top-p sampling \citep[aka nucleus sampling;][]{2019arXiv190409751H} approach is followed (after the grammar constrains have been applied in the case of the highlights subtask), where only the top tokens will be considered (up to a cumulative score of 0.95), and a temperature of 0.7 to favor precision over creativity (low values makes top tokens more likely) and a repeat penalty of 1.1 is used to prevent loops.

Thanks to the workshop organizers, the generated text highlights and summaries that support the preassigned suicidal risk level were automatically evaluated against a test set annotated by domain experts (who manually generated gold highlights and summaries). The computed metrics to evaluate highlights are:

\begin{itemize}
    \item Recall: For every gold highlight, find the generated highlight with the highest semantic similarity \citep[based on BERTScore;][]{2019arXiv190409675Z} and compute the average across users. It measures how relevant the highlights are as supporting evidence.
    \item Precision: For every generated highlight, find the gold highlight with the highest semantic similarity and compute the average across users. It measures the quality of the generated highlights.
    \item Weighted Recall: Sum the gold and generated highlights lengths (i.e., number of tokens) per user. If the generated length is greater than the gold one, correct the calculated recall value $R$ with the length ratio: $R_{\text{weighted}} = R \times \frac{L_{\text{gold}}}{L_{\text{candidate}}}$. It measures how relevant the highlights are as supporting evidence and if lengths are similar to human-highlighted ones.
    \item Harmonic Mean: Balances between precision and recall (the unweighted version).
\end{itemize}

Regarding the evaluation of summaries, the computed metrics are:

\begin{itemize}
    \item Consistency: Using a natural language inference (NLI) model, obtain the probability $p$ of each generated sentence (hypothesis) contradicting the gold sentence (premise). Then average $1 - p$ across all sentences and users. It measures lack of contradiction.
    \item Contradiction: Similar to the previous one, but directly takes the maximum contradiction probability and averages all sentences and users. Hence, it penalizes information that contradicts the gold summary, and lower scores are better.
\end{itemize}

\section{Results}

The CLPsych 2024 shared task only accepted three submissions per team, but the organizers were kind enough to evaluate additional submissions that are not considered for the workshop ranking. For the competition, I submitted the output from OpenHermes, Orca 2, and Starling. Orca 2 was selected as it is the sole model based on Llama 2, while the other two were chosen for their standings in the LLM leaderboards. In the final official ranking, OpenHermes produced the best results. For highlights, based on recall and harmonic mean metrics, it ended in the modest 10th position (out of 15). However, if the weighted recall were considered instead, it would have ended in the 3rd position. This shows that OpenHermes' length of its generated highlights are closer to human-highlighted ones compared to other systems. Regarding summaries, based on the consistency metrics, OpenHermes ended in an outstanding 2nd position (out of 14). If the organizer would have considered the contradiction metrics, then it would have fallen to a (still honorable) 3rd position. 

\begin{table*}
\centering
{\small
\begin{tabular}{l|cccc|cc}
\hline
 & \multicolumn{4}{c|}{\textbf{Highlights}}  & \multicolumn{2}{c}{\textbf{Summaries}} \\ \hline
\textbf{Model} & \textbf{Recall} & \textbf{Precision} & \textbf{Weighted Recall} & \textbf{Harmonic Mean} & \textbf{Consistency} & \textbf{Contradiction} \\ \hline
OpenHermes & 0.907 & 0.912 & 0.738 & 0.909 & 0.976 & \textbf{0.079} \\
Orca 2 & 0.904 & \textbf{0.914} & 0.777 & 0.909 & 0.971 & 0.104 \\
Starling & 0.907 & 0.913 & 0.766 & 0.910 & \textbf{0.977} & 0.083 \\
\hline
Dolphin & \textbf{0.910} & 0.913 & 0.736 & \textbf{0.911} & 0.971 & 0.093 \\
Mistral & 0.902 & 0.913 & 0.799 & 0.907 & 0.969 & 0.105 \\
Zephyr & 0.894 & \textbf{0.914} & \textbf{0.803} & 0.903 & 0.974 & 0.085 \\ \hline
\end{tabular}
}
\caption{Performance metrics for all the evaluated models. The last three models did not enter the CLPsych 2024 shared task competition. The best scores per metric are highlighted in bold.}
\label{tab:metrics}
\end{table*}

Beyond the workshop competition, and in the interest of better assessing all the considered LLMs, the performance metrics for all the evaluated models can be found in Table~\ref{tab:metrics}.
For the highlights, the best performing models are Dolphin and Zephyr, depending if we consider the weighted or unweighted recall metrics. Zephyr produces highlights of a length that is more similar to the human-made highlights, but Dolphin generates highlights that are more relevant. Regarding summaries, OpenHermes and Starling are in the lead, depending if we give a higher importance to being consistent or minimizing contradictions. OpenHermes generates summaries with the lowest level of contradiction, and its consistency level is only slightly below Starling, hence it would be fair to claim that it is the best model for this subtask.

It is also worth exploring the evaluation metrics split by the preassigned suicidal risk level (see Table~\ref{tab:highlights_metrics_per_risk_level} and Table~\ref{tab:summary_metrics_per_risk_level}). There is no single model that excels at all risk levels, suggesting that a combined strategy could lead to even better overall results. Additionally, almost for all models and metrics, the performance correlates with the suicidal risk level: the higher the risk, the better the performance of the model.

\begin{table*}
\centering
{\small
\begin{tabular}{l|rrr|rrr|rrr|rrr}
\hline
 & \multicolumn{3}{c|}{\textbf{Recall}} & \multicolumn{3}{c|}{\textbf{Precision}} & \multicolumn{3}{c|}{\textbf{Weighted Recall}} & \multicolumn{3}{c}{\textbf{Harmonic Mean}} \\
\hline
\textbf{Model / Risk} & \textbf{Low} & \textbf{Mod.} & \textbf{High} & \textbf{Low} & \textbf{Mod.} & \textbf{High} & \textbf{Low} & \textbf{Mod.} & \textbf{High} & \textbf{Low} & \textbf{Mod.} & \textbf{High} \\
\hline
OpenHermes & 0.900 & 0.904 & \textbf{0.915} & 0.896 & 0.909 & 0.922 & 0.677 & 0.739 & 0.759 & 0.898 & 0.906 & \textbf{0.919} \\
Orca 2 & 0.902 & 0.903 & 0.905 & \textbf{0.907} & \textbf{0.914} & 0.919 & 0.723 & 0.785 & 0.778 & \textbf{0.905} & 0.908 & 0.911 \\
Starling & 0.892 & 0.909 & 0.907 & 0.893 & 0.911 & 0.924 & 0.705 & 0.763 & 0.794 & 0.892 & 0.910 & 0.915 \\
\hline
Dolphin & 0.901 & \textbf{0.912} & 0.912 & 0.894 & 0.913 & 0.920 & 0.632 & 0.748 & 0.750 & 0.897 & \textbf{0.912} & 0.915 \\
Mistral & \textbf{0.905} & 0.898 & 0.909 & 0.898 & 0.910 & \textbf{0.925} & 0.658 & \textbf{0.816} & \textbf{0.813} & 0.901 & 0.904 & 0.917 \\
Zephyr & 0.890 & 0.893 & 0.896 & 0.900 & \textbf{0.914} & 0.917 & \textbf{0.792} & 0.811 & 0.791 & 0.895 & 0.903 & 0.906 \\
\hline
\end{tabular}
}
\caption{Performance metrics for the highlights subtask, split by users with different suicidal risk level (low, moderate, or high). The best scores per metric and risk level are highlighted in bold.}
\label{tab:highlights_metrics_per_risk_level}
\end{table*}

\begin{table}
\centering
{\small
\begin{tabular}{l@{\hspace{5pt}}|@{\hspace{5pt}}r@{\hspace{5pt}}r@{\hspace{5pt}}r@{\hspace{5pt}}|@{\hspace{5pt}}r@{\hspace{5pt}}r@{\hspace{5pt}}r}
\hline
\multicolumn{1}{l}{} & \multicolumn{3}{c}{\textbf{Consistency}} & \multicolumn{3}{c}{\textbf{Contradiction}} \\
\hline
\textbf{Model / Risk} & \textbf{Low} & \textbf{Mod.} & \textbf{High} & \textbf{Low} & \textbf{Mod.} & \textbf{High} \\
\hline
OpenHermes & 0.937 & 0.977 & \textbf{0.986} & 0.178 & \textbf{0.078} & \textbf{0.045} \\
Orca 2 & 0.958 & 0.975 & 0.970 & 0.125 & 0.092 & 0.119 \\
Starling & \textbf{0.962} & \textbf{0.978} & 0.979 & \textbf{0.113} & 0.079 & 0.079 \\
\hline
Dolphin & 0.948 & 0.973 & 0.975 & 0.165 & 0.084 & 0.087 \\
Mistral & 0.931 & 0.976 & 0.968 & 0.205 & 0.084 & 0.110 \\
Zephyr & 0.944 & 0.976 & 0.981 & 0.161 & 0.081 & 0.068 \\
\hline
\end{tabular}
}
\caption{Performance metrics for the summarization subtask, split by users with different suicidal risk level (low, moderate, or high). The best scores per metric and risk level are highlighted in bold.}
\label{tab:summary_metrics_per_risk_level}
\end{table}

Finally, in terms of computation, the average inference time was of 40 minutes to extract highlights from 162 posts ($\sim$14.8 seconds per post), and 30 minutes to generate summaries for 125 users ($\sim$14.4 seconds per user). These are extremely competitive numbers for LLMs running on a consumer-grade machine.

\section{Discussion}

The OpenHermes' generated highlights and summaries, when compared to other submitted systems to the CLPsych 2024 shared task competition, ended up with remarkable comparative metrics for an approach that has used cost-effective "open-source" LLMs without any specific fine-tuning for these specific tasks. The highlights subtask seems to be the one with more margin of improvement, especially if we only consider the unweighted recall (where matching highlight lengths are not taken into account). It would have been interesting to make a manual human-based evaluation, comparing the generated highlights with the golden ones (which has not been released publicly), to better understand the discrepancy between the unweighted and weighted recall metrics (10th vs 3rd position in the final ranking, respectively) and justify selecting one over the other. In any case, these extraordinary results seem to signal the potential that this approach may have at other relatively similar tasks such as Named Entity Recognition. Regarding the generation of summaries, both evaluation metrics placed this approach in the top 3 ranking, a stunning result for a model that has not been trained specifically for psychological assessments.

OpenHermes seems to be the best well-balance model and one of the best for summarization, but if we consider all the evaluated LLMs, Dolphin and Zephyr perform better in the highlights subtask. However, these results would likely change if other prompt templates were used. For instance, the crafted prompt includes only one example with a high suicidal risk level, and we observed that almost all models perform better for comments from high risk users. Expanding the prompt to include more examples of different risk levels could potentially improve the overall performance.

\section{Conclusion}

Six different "open-source" Large Language Models were evaluated to accomplish the shared task proposed by the CLPsych 2024 workshop. This work demonstrated that following a relatively simple approach, mainly consisting on a well structured prompt with one single example, can be used with cost-effective LLMs to extract highlights and generate comprehensive and consistent summaries that justify a preassigned suicidal risk level of users who participate in online text-based forums. This approach does not rely on complex operations such as further training or fine-tuning the models to adapt them to the goal in hand. Hence, existing "open-source" models with moderate hardware requirements can successfully run locally to support psychological assessments. This approach facilitates respecting privacy rules, best ethical practices and other local, national, and international regulations.

\section{Limitations}

This study has considered a selection of six models based on two existing public rankings, but there are many more "open-source" LLMs available. In particular, there are models with even larger number of parameters that could still be run in advanced commodity hardware such as gaming computers. It would have been interesting to evaluate how differently these larger models perform compared to the selected 7 billion parameter models. The presented approach is also highly dependent on the prompt template, this work has not explored and compared other prompts with more embedded examples or different prompt wording.

It is important to recognize that LLMs are trained on datasets that are not necessarily representative of the global population (e.g., many biases may be present), and they may not always accurately interpret the nuances of human psychology and emotions, leading to potential misinterpretation with severe implications. Hence, using LLMs for psychological assessments has associated risks that need to be properly handled and mitigated.

Finally, not having direct access to the golden highlights and summaries makes understanding certain results a bit more difficult. 

\section{Ethics}

Secure access to the shared task dataset was provided with IRB approval under University of Maryland, College Park protocol 1642625 and approval by the Biomedical and Scientific Research Ethics Committee (BSREC) at the University of Warwick (ethical application reference BSREC 40/19-20). The author\footnote{\url{https://blancocuaresma.com/s/}} signed a data use agreement that establishes multiple requirements such as the strict use of the dataset for only the CLPsych shared task, protecting its content and the users privacy.

\section{Acknowledgements}

The author is particularly grateful to the anonymous users of Reddit whose data feature in this year’s shared task dataset, the American Association of Suicidology for making the dataset available, the individuals involved in manually annotating the dataset, and to the organizers of the CLPsych 2024 workshop/shared task. This work could not have been possible without the support of NASA Science Explorer\footnote{\url{https://scixplorer.org/}} and its brilliant team at the Harvard-Smithsonian Center for Astrophysics (USA), as well as Dr. Marzia De Lucia's support at the Lausanne University Hospital/University of Lausanne (Switzerland).

\bibliography{custom}

\appendix

\section{Prompt for Highlights Extraction}
\label{sec:appendix_prompt_highlights}

The specific text included as an example in the following prompt has been altered and paraphrased to safeguard user privacy (the actual prompt contained a real example):

\fbox{
  \begin{minipage}{0.9\columnwidth}
      \tiny
      \raggedright
      \texttt{\\<|im\_start|>system\\
As a psychologist specializing in suicidal ideation, you are adept at identifying key evidences within brief online text comments that assess an individual's risk of suicide. These evidences encompass various aspects: Emotions (the individual's feelings), Cognitions (thoughts and perceptions about suicide), Behavior and Motivation (actions or behaviors linked to suicide), Interpersonal and Social Support (availability of social support and stable relationships), Mental Health-Related Issues (potential psychiatric diagnoses), and Risk Factors (including socioeconomic and demographic aspects). The risk levels are categorized as low (indicating minimal perceived risk of suicide), moderate (suggesting a tangible risk of a suicide attempt), and high (implying a strong likelihood of a suicide attempt in the near future). You will now analyze several texts to extract pertinent excerpts as evidences that effectively evaluate the individual's risk of suicide.<|im\_end|>\\
<|im\_start|>user\\
Risk level: high\\
Text: Recently, I attempted suicide by consuming an unspecified amount of prescription medications. After realizing that the pills were not lethal, I desperately prayed for an end to my suffering, wishing to escape the torment inflicted by those who profess their love yet demonstrate indifference through their actions.<|im\_end|>\\
<|im\_start|>assistant\\
}
[
\texttt{"Recently, I attempted suicide", "I desperately prayed for an end", "those who profess their love yet demonstrate indifference through their actions"}
]
\texttt{\\<|im\_start|>user\\
Risk level: {risk\_level}\\
Text: {user\_comment}\\
<|im\_start|>assistant\\
}
  \end{minipage}
}

The prompt above follows the Chat Markup Language (ChatML) format, which is used by OpenHermes, Orca 2, and Dolphin. The prompt format was adapted to follow the appropriate one for the rest of the models (e.g., Starling uses \textit{GPT4 User:}, \textit{GPT4 Assistant}, and no tag for the system message).

\section{Grammar for Highlights Extraction}
\label{sec:appendix_grammar_highlights}

Simplified example of a grammar used to extract highlights from the following text: \textit{'Recently, I attempted suicide by consuming an unspecified amount of prescription medications.'}.

\fbox{
  \begin{minipage}{0.9\columnwidth}
      \tiny
      \raggedright
      \texttt{\\root ::= "[" h (", " h)* "]"\\
h ::= "\"" (t0 | t1 | t2 | t3 | t4 | t5 | t6 | t7 | t8 | t9 | t10 | t11) "\""\\
t0 ::= "Recently," (" " t1)?\\
t1 ::= "I" (" " t2)?\\
t2 ::= "attempted" (" " t3)?\\
t3 ::= "suicide" (" " t4)?\\
t4 ::= "by" (" " t5)?\\
t5 ::= "consuming" (" " t6)?\\
t6 ::= "an" (" " t7)?\\
t7 ::= "unspecified" (" " t8)?\\
t8 ::= "amount" (" " t9)?\\
t9 ::= "of" (" " t10)?\\
t10 ::= "prescription" (" " t11)?\\
t11 ::= "medications." (" " t12)?\\
t12 ::= ""\\
}
  \end{minipage}
}

A response that respects this grammar would be: \textit{'["I attempted suicide", "prescription medication"]'}. However, the grammar does not prevent the model from extracting repetitive strings or out of order strings (e.g., \textit{'["suicide", "I attempted suicide", "medications.", "I attempted suicide"]'}), but this behavior was not observed (probably thanks to the repeat penalty).

\section{Prompt for Summarization}
\label{sec:appendix_prompt_summary}

As in Appendix~\ref{sec:appendix_prompt_highlights}, the specific text included as an example in the following prompt was altered and paraphrased to safeguard user privacy:

\fbox{
  \begin{minipage}{0.9\columnwidth}
      \tiny
      \raggedright
      \texttt{\\<|im\_start|>system\\
As a psychologist specializing in suicidal ideation, you excel in crafting concise summaries that collate evidences supporting a designated suicide risk level for online texts. These evidences encompass various aspects: Emotions (the individual's feelings), Cognitions (thoughts and perceptions about suicide), Behavior and Motivation (actions or behaviors linked to suicide), Interpersonal and Social Support (availability of social support and stable relationships), Mental Health-Related Issues (potential psychiatric diagnoses), and Risk Factors (including socioeconomic and demographic aspects). The risk levels are categorized as low (indicating minimal perceived risk of suicide), moderate (suggesting a tangible risk of a suicide attempt), and high (implying a strong likelihood of a suicide attempt in the near future). You will now analyze various texts and succinctly summarize the evidence that substantiates the assigned risk level for each case.<|im\_end|>\\
<|im\_start|>user\\
Risk level: high\\
Text: Recently, I attempted suicide by consuming an unspecified amount of prescription medications. After realizing that the pills were not lethal, I desperately prayed for an end to my suffering, wishing to escape the torment inflicted by those who profess their love yet demonstrate indifference through their actions. I am prone to anxiety, and for the past two weeks, I've been coerced into tolerating the intrusive presence of my housemate's girlfriend. Despite my patience, my attempt to diplomatically express the need for boundaries was met with coercion, exacerbating my sense of violation.<|im\_end|>\\
<|im\_start|>assistant\\
This person is at high risk because they describe a recent suicide attempt. They express a wish to be dead, extreme hopelessness, and a sense of feeling trapped. Their overall tone is aroused and agitated. They feel disconnected from others, and bullied by others. They experience extreme anxiety.<|im\_end|>\\
<|im\_start|>user\\
Risk level: {risk\_level}\\
Text: {user\_comments}<|im\_end|>\\
<|im\_start|>assistant\\
This person is at {risk\_level} risk.
}
  \end{minipage}
}

\end{document}